\documentclass[10pt,twocolumn,letterpaper]{article}

\usepackage{cvpr}              

\usepackage{graphicx}
\usepackage{amsmath}
\usepackage{amssymb}
\usepackage{booktabs}
\usepackage[utf8]{inputenc}
\usepackage{algorithm}
\usepackage{algpseudocode}

\usepackage[pagebackref,breaklinks,colorlinks]{hyperref}

\usepackage[capitalize]{cleveref}
\crefname{section}{Sec.}{Secs.}
\Crefname{section}{Section}{Sections}
\Crefname{table}{Table}{Tables}
\crefname{table}{Tab.}{Tabs.}

\def\cvprPaperID{*****} 
\def\confName{CVPR}
\def\confYear{2024}

\begin{document}

\title{Low-Light Image Enhancement Framework for Improved Object Detection in Fisheye Lens Datasets}


\author{Dai Quoc Tran \thanks{Equal contribution}\\
Global Frontiers of Resilient EcoSmart City,\\
Sungkyunkwan University, South Korea\\
{\tt\small daitran@skku.edu}
\and
Armstrong Aboah *\\
Depart. of Civil, Construction \& Env. Engineering\\
North Dakota State University, United States\\
{\tt\small armstrong.aboah@ndsu.edu}
\and
Yuntae Jeon\\
Department of Global Smart City\\
Sungkyunkwan University, South Korea\\
{\tt\small jyt0131@g.skku.edu}
\and
Maged Shoman\\
Department of Civil and Env Engineering\\
University of Central Florida, United States \\
{\tt\small ma657478@ucf.edu}
\and
Minsoo Park\\
Sungkyun AI Research Institute\\
Sungkyunkwan University, South Korea\\
{\tt\small pms5343@skku.edu}
\and
Seunghee Park \thanks{Corresponding author}\\
School of Civil, Arch Eng. and Landscape Architecture\\
Sungkyunkwan University, South Korea\\
{\tt\small shparkpc@skku.edu}
}

\maketitle

\begin{abstract}
This study addresses the evolving challenges in urban traffic monitoring detection systems based on fisheye lens cameras by proposing a framework that improves the efficacy and accuracy of these systems. In the context of urban infrastructure and transportation management, advanced traffic monitoring systems have become critical for managing the complexities of urbanization and increasing vehicle density. Traditional monitoring methods, which rely on static cameras with narrow fields of view, are ineffective in dynamic urban environments, necessitating the installation of multiple cameras, which raises costs. Fisheye lenses, which were recently introduced, provide wide and omnidirectional coverage in a single frame, making them a transformative solution. However, issues such as distorted views and blurriness arise, preventing accurate object detection on these images. Motivated by these challenges, this study proposes a novel approach that combines a \textbf{\textit{transformer-based image enhancement framework}} and \textbf{\textit{ensemble learning technique}} to address these challenges and improve traffic monitoring accuracy, making significant contributions to the future of intelligent traffic management systems. Our proposed methodological framework won \textbf{5th place} in the 2024 AI City Challenge, Track 4, with an F1 score of \textbf{0.5965} on experimental validation data. The experimental results demonstrate the effectiveness, efficiency, and robustness of the proposed system. Our code is publicly available at \href{https://github.com/daitranskku/AIC2024-TRACK4-TEAM15}{https://github.com/daitranskku/AIC2024-TRACK4-TEAM15}.

\end{abstract}

\section{Introduction}
\label{sec:intro}

In the field of urban infrastructure and transportation management, the development of advanced traffic monitoring systems has become a crucial solution to the growing challenges posed by urbanization and increasing vehicular density\cite{al2021improved,mandal2020artificial,chu2022nafssr,shoman2022multi}. These systems, utilizing state-of-the-art technologies such as computer vision, machine learning, and data analytics, are tasked with ensuring not only the smooth flow of traffic but also enhancing safety\cite{aboah2021vision, 9801825, jeon2023leveraging, tran2020damage} and efficiency on busy roadways\cite{sarrab2020development}. As cities grow in size and population, there is an increasing need for advanced traffic monitoring and management solutions that surpass traditional strategies.


Traditional traffic monitoring methods, which rely on static cameras with limited fields of view (FoV) \cite{fedorov2019traffic,madhavi2023traffic,yadav2023video}, have proven insufficient in dealing with the dynamic and complex nature of modern urban environments. These cameras typically provide narrow perspectives of roadways and intersections, requiring the deployment of multiple cameras to achieve comprehensive coverage\cite{gochoo2023fisheye8k}. This not only raises the cost and complexity of surveillance infrastructure but also creates blind spots and gaps in monitoring, particularly in areas with complex road layouts or high traffic volumes. The need for real-time insights, comprehensive coverage, and adaptive response mechanisms has highlighted the importance of more advanced and versatile surveillance techniques.

In recent years, the introduction of fisheye lenses has revolutionized surveillance and traffic monitoring systems due to their ability to provide natural, wide, and omnidirectional coverage\cite{gochoo2023fisheye8k}. This unique feature addresses a significant limitation of traditional cameras with narrow fields of view (FoV), allowing for the capture of large scenes in a single frame—an accomplishment not possible with conventional counterparts. Fisheye lenses in traffic monitoring systems have proven particularly advantageous in reducing the number of required cameras, offering a cost-effective solution to cover broader views of streets and intersections \cite{ardianto2023fast}. \textit{However, this innovation comes with its own set of challenges, as fisheye cameras inherently present distorted views, which require sophisticated design approaches for image undistortion and unwarping \cite{gochoo2023fisheye8k}.  Additionally, objects at the edges or far ends of the captured scenes appear small and blurry. This makes it difficulty for object detection systems to accurately identify important elements such as cars, pedestrians, and road signs during traffic monitoring\cite{gochoo2023fisheye8k}}. These challenges underscore the need for dedicated strategies to address distortions and blurriness during image processing, and that is what this study seeks to do.


Inspired by these challenges, the overarching goal of this study is to develop a robust framework for traffic monitoring using data from fisheye lens cameras. To achieve this goal, we propose a \textbf{\textit{Low-Light Image Enhancement Framework}} to enhance image quality, resulting in improved object detection accuracy for fisheye images. The proposed image enhancement framework aims to \textit{improve image clarity and accuracy in object detection by addressing poor visibility at night and blurriness in video-generated images}. To achieve a robust objection detection model, the study incorporates the principle of \textbf{\textit{ensemble learning}}, drawing upon diverse state-of-the-art object detection models for this task. By using the ensemble learning technique, we mitigate the limitations associated with using individual models for object detection tasks. The models utilized in this study include collaborative detection transformer (Co-DETR), You Only Look Once (YOLOv8x), and YOLOv9.

To this end, the study's main contributions can be summarized as follows:
\begin{enumerate}
    \item We propose a unique data preprocessing framework called the \textbf{\textit{Low-Light Image Enhancement Framework}}. This framework utilizes a transformer-based image enhancement technique, NAFNET \cite{chu2022nafssr}, to improve image clarity by removing blurriness, and GSAD \cite{hou2024global} to convert nighttime images (low illumination) to daytime images (high illumination) to improve object detection accuracy in fisheye images during model training. To enhance object detection accuracy during inference, the study used a super-resolution postprocessing technique to increase image pixels, as well as an \textbf{\textit{ensemble model technique}} for robust detection.
    \item We performed a detailed comparative analysis of our proposed ensembled model to other state-of-the-art object detection models (Co-DETR, YOLOv8x, and YOLOv9e). By evaluating their performance in detecting objects from fisheye lens-captured images, we aimed to demonstrate the superiority of our proposed model over the current state-of-the-art models. In addition, we demonstrate that our pre- and post-processing techniques are effective in leading to improved object detection. 
    
    \item Our proposed approach showed its robustness in AICity Challenge Track 4, placing \textbf{5th out of 52 teams}.
\end{enumerate}

The experimental results of this study hold paramount importance in shaping the future of intelligent traffic monitoring systems, particularly those utilizing fisheye lens cameras. The proposed robust framework, anchored by the transformer-based image enhancement technique and enriched by ensemble learning, represents a significant stride towards overcoming the challenges posed by fisheye distortions in urban environments. These results offer valuable insights into the feasibility and real-world applicability of our approach, providing a tangible foundation for the advancement of traffic monitoring technology.

The remainder of the paper is structured as follows: In section 2, we present a discussion of related works. Section 3 discusses our methodological framework. In section 4, we present our data and experimental findings, which demonstrate the efficacy of our proposed method in detecting objects in fisheye lens cameras. In section 5, we discuss the implications of our findings and make suggestions for future research in this field.

\section{Related Work}
\label{sec:related}
Traffic surveillance has advanced significantly in recent years as a result of the convergence of computer vision, machine learning, and data analytics. Our ability to accurately detect and track vehicles, pedestrians, and motorists in surveillance videos is critical for ensuring road safety, optimizing traffic flow, and improving overall transportation efficiency. Object detection, a critical task in traffic surveillance systems, has advanced dramatically as new algorithms and techniques emerge. This literature review focuses on three techniques or algorithms of object detection: multiple/two-stage detectors, single-stage detectors,  and transformer-based models. 

\subsection{Multiple/Two-stage detectors}

A variety of studies have investigated the utilization of two-stage detection algorithms in transportation systems\cite{shoman,shirpour2021traffic,wang2014hybrid}.  Shirpour et al. developed a real-time traffic object detection system, achieving 91\% accuracy by employing a combination of multi-scale HOG-SVM and Faster R-CNN models\cite{shirpour2021traffic}.  Also, Nizar et al. utilized HOG and SVM for feature extraction and KLT for object counting, achieving an average accuracy of 95.15\% [18].  Wang \& Zhang proposed a hybrid method for vehicle detection, integrating shadow area search with ROI, HOG, and SVM algorithms, along with K-means clustering\cite{wang2014hybrid}.  Gavrila introduced a two-step approach for pedestrian detection, leveraging contour features and hierarchical template matching in the first step, and intensity features and pattern classification in the second step\cite{gavrila2000pedestrian}. Additionally,  Zhang proposed a vision-based method for vehicle detection, featuring an improved common region algorithm for background subtraction and a threshold segmentation method for object extraction, achieving enhanced accuracy and stability compared to existing algorithms\cite{wang2010boosting}. Collectively, these studies underscore the potential of two-stage detection algorithms in enhancing the accuracy and robustness of object detection in transportation systems.

\subsection{Single-stage detectors}
Recent advancements in object detection have witnessed the emergence of single-stage detection algorithms, offering simpler and faster alternatives to traditional two-stage methods.  Ye et al introduced the feature-enhanced single-shot detector (FE-SSD) for railway traffic, significantly improving feature discrimination and robustness\cite{ye2020autonomous}.  Alvarez et al proposed a monocular target detection system for transport infrastructures, incorporating vanishing point extraction for automatic camera calibration and a background subtraction method for object segmentation [7].  Stuparu et al presented a one-stage object detection model for vehicle detection in overhead satellite images, achieving high accuracy and low detection time\cite{stuparu2020vehicle}.  Qiu et al., further enhanced vehicle detection in intelligent transportation systems with a deep learning-based algorithm, achieving a 99.82\% recognition rate in real traffic scenes\cite{qiu2021deep}. Redmon et al. introduced YOLO, a novel approach to object detection that significantly revolutionizes single-stage object detection frameworks\cite{redmon2016you,aboah2023real,shoman2022region,agorku2023real,soltanikazemi2023real,aboah2023deepsegmenter}. YOLO utilizes a single neural network to directly forecast bounding boxes and class probabilities from complete images in one assessment, enabling end-to-end optimization for detection efficacy.

\subsection{Transformer-based detectors}

Transformers have recently emerged as a significant advancement in computer vision, particularly in the realm of object detection. These models have introduced end-to-end learning systems and have been integrated into various architectures to enhance detection performance. Recent advancements in object detection have witnessed the emergence of transformer-based algorithms, which have demonstrated promising results in enhancing both accuracy and convergence time\cite{vaidwan2021study}. A comprehensive review of object detection algorithms, encompassing transformer-based detectors, has highlighted substantial progress in the field, particularly in the era of deep learning \cite{song2022extendable}. The integration of Vision and Detection Transformers (ViDT) has further enhanced the efficiency and effectiveness of object detection, with ViDT+ achieving high scalability for large models\cite{redmon2016you}.   Carion et al present a novel approach named DEtection TRansformer (DETR), which conceptualizes object detection as a direct set prediction problem. DETR simplifies the detection pipeline by eliminating the necessity for hand-designed components such as non-maximum suppression or anchor generation\cite{nguyen2014object}. Shou et al used the MS Transformer model to enhance object detection in medical images by addressing challenges such as low resolution, high noise, and small object size\cite{shou2022object}. It surpasses existing methods on benchmark datasets like DeepLesion and BCDD, demonstrating superior performance in medical image analysis.  extend ViDT to ViDT+ to facilitate joint-task learning for object detection and instance segmentation \cite{shou2022object}.

\section{Methodology}
\label{sec:method}

\begin{figure*}[ht]
  \centering  
  \includegraphics[width=\linewidth]{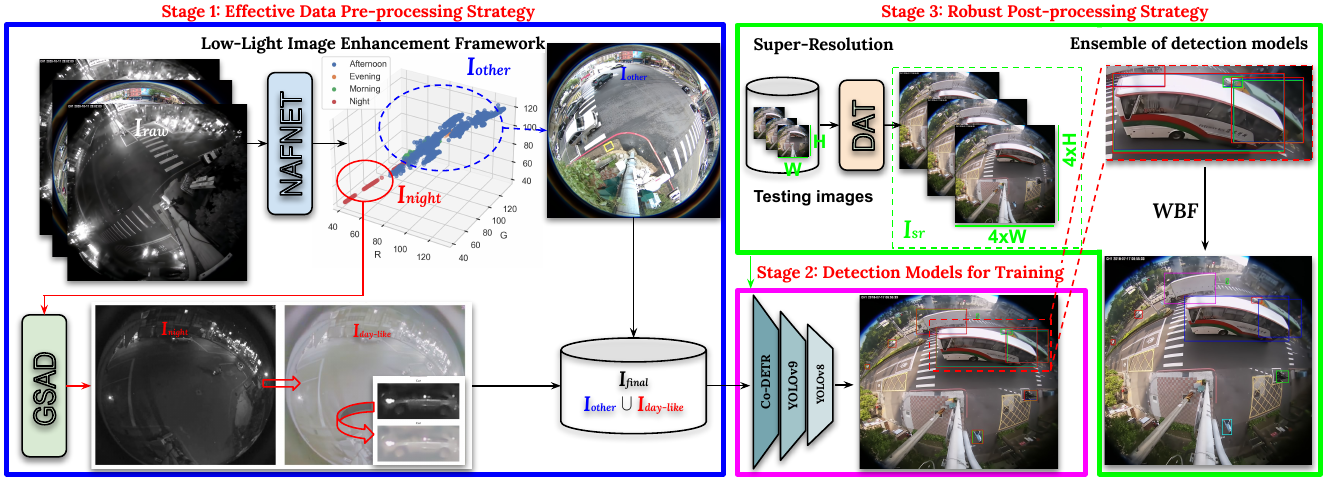}
  \caption{The proposed approach has three stages. 1) effective data preprocessing strategy (NAFNET \cite{chen2022simple}; GSAD \cite{hou2024global}), 2) detection models for training ( Co-DETR \cite{zong2023detrs};YOLOv8\cite{aboah2023real};YOLOv9\cite{wang2024yolov9} ), and 3) robust post-processing strategy (DAT \cite{chen2023dual}; WBF \cite{solovyev2021weighted}) }
  \label{fig:proposedapproach}
\end{figure*}

Figure \ref{fig:proposedapproach} and Algorithm \ref{alg:object_detection} illustrated our proposed approach. Our approach consists of three major stages: 1) an effective data pre-processing step, 2) model training, and 3) a robust post-processing strategy.


\subsection{Effective Data Pre-processing Strategy}
In order to address the data challenges such as blurriness, small object detection at image edges, and low illumination in nighttime images, we proposed an effective preprocessing strategy called \textbf{\textit{Low-Light Image Enhancement Framework}}. The proposed framework consists of 3 key steps, namely image enhancement, clustering by illumination, and night-to-day image conversion. 

\begin{figure}[H]
  \centering
    \includegraphics[width=8cm]{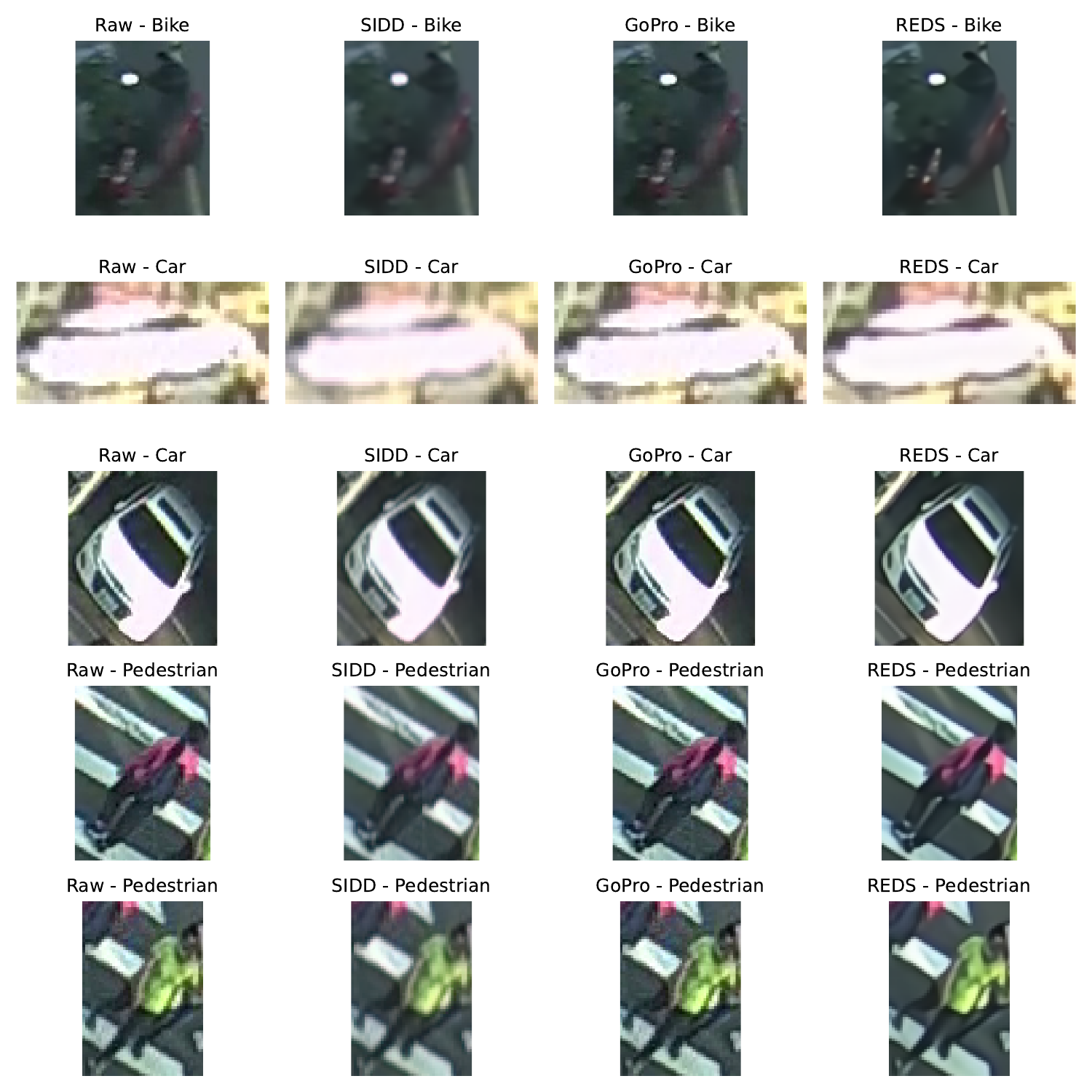}
    \caption{Compare NAFNet \cite{chen2022simple} image enhancement algorithm with different trained dataset. }
    \label{figenhanced_1}
\end{figure}

\vspace{0.1in}
\textbf{NAFNet-based Image Enhancement.} The initial phase involves enhancing the raw fisheye images ($\mathcal{I}_{raw}$) to rectify issues related to low light and noise, common in such datasets. Utilizing the NAFNet image enhancement model, each raw image $i$ is processed to yield an enhanced version $E(i)$, forming a set of enhanced images, $\mathcal{I}_{enh}$. This step not only improves the visual quality of the images but also prepares them for more accurate object detection by reducing noise and enhancing details. As shown in Figure \ref{figenhanced_1}, by using NAFNet\cite{chen2022simple}, objects are enhanced and show clearer quality compared to the original image. 

To achieve the best image enhancement on our training images, three pre-trained NAFNet models were considered, as illustrated in Figure \ref{figenhanced_1}. They are NAFNet trained on SIDD\cite{abdelhamed2018high}, GoPro\cite{nah2017deep}, and REDS\cite{nah2019ntire}. Each pre-train model demonstrated the capacity of the NAFNet algorithm to improve image clarity, sharpness, and overall visibility of details. For example, the Bike images in Figure \ref{figenhanced_1} show less noise and clearer contours after enhancement. The Car images also show more defined shapes and textures, while the Pedestrian images show more details and contrast, making features easier to see. In this research, NAFNet network that was pre-trained on REDS was utilized for improving image quality.

\begin{figure}[ht]
  \centering
  \includegraphics[width=0.7\linewidth]{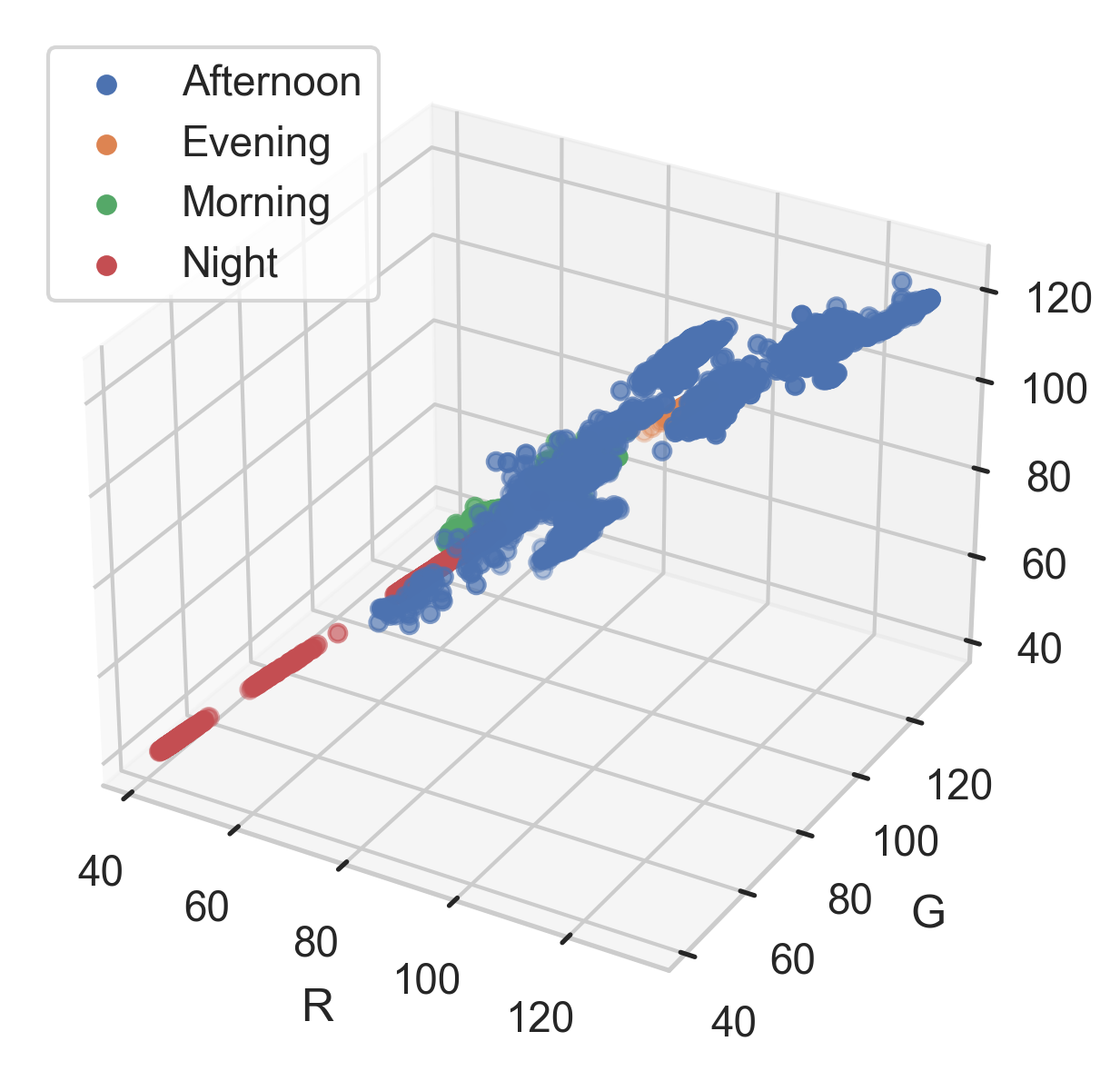}
   \caption{Compare mean value from all scenario images.}
   \label{fig:mean_night_value}
\end{figure}

\vspace{0.1in}
\vspace{0.1in}
\vspace{0.1in}
\textbf{Clustering by illumination condition.} Post-enhancement, the images are clustered into two subsets based on their mean values, distinguishing between night-time and other time-of-the-day images. This step is crucial for tailoring subsequent processing to the specific challenges associated with low-light conditions. Night-time images, $\mathcal{I}_{night}$, are identified through a predefined threshold, $\mathcal{T}_{night}$, whereas the remaining images are categorized as $\mathcal{I}_{other}$. Figure \ref{fig:mean_night_value} provided 3D scatter plot of the mean color values in RGB color space under four different illumination conditions of the FishEye8K dataset: Morning, Afternoon, Evening, and Night. Clearly, there is a distinct shift towards the lower end of the RGB value spectrum from Night condition.

\vspace{0.1in}
\vspace{0.1in}

\begin{figure}[H]
  \centering
  \includegraphics[width=\linewidth]{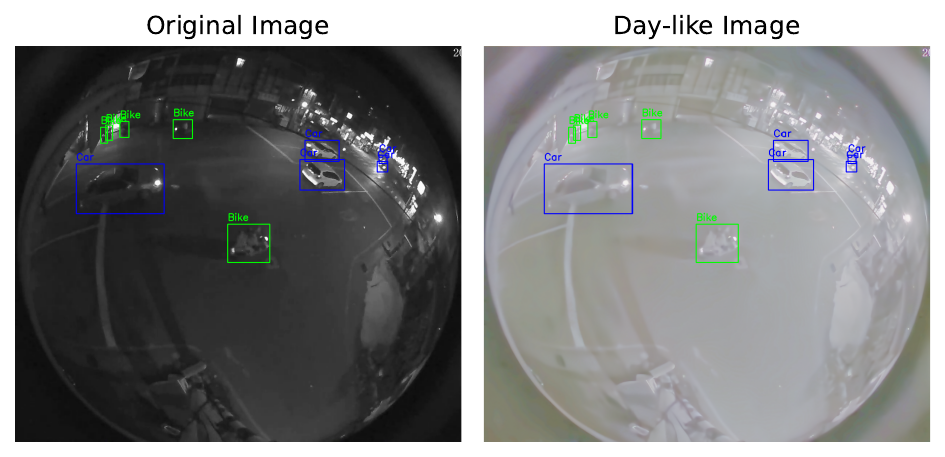}
   \caption{Converting night time image to day-like image using GSAD model.}
   \label{fig:lolv2_compare}
\end{figure}

\begin{algorithm}[H]
\caption{Object Detection in Fisheye Datasets}\label{alg:object_detection}
\begin{algorithmic}[1]
\State \textbf{Input:} $\mathcal{I}_{raw}$ - Set of raw fisheye images.
\State \textbf{Output:} $\mathcal{O}$ - Object detection results.

\Statex
\Procedure{EnhanceImages}{$\mathcal{I}_{raw}$}
    \State Initialize $\mathcal{I}_{enh}$ as an empty set
    \For{each image $i$ in $\mathcal{I}_{raw}$}
        \State $E(i) \gets$ Apply enhancement model $E$ to $i$
        \State Add $E(i)$ to $\mathcal{I}_{enh}$
    \EndFor
\EndProcedure

\Statex
\Procedure{ClusteringByIlluminationCondition}{$\mathcal{I}_{enh}$}
    \State Initialize $\mathcal{I}_{night}$, $\mathcal{I}_{other}$ as empty sets
    \For{each image $i$ in $\mathcal{I}_{enh}$}
        \If{$L(i) < T_{night}$}
            \State Add $i$ to $\mathcal{I}_{night}$
        \Else
            \State Add $i$ to $\mathcal{I}_{other}$
        \EndIf
    \EndFor
\EndProcedure

\Statex
\Procedure{ConvertNightToDay}{$\mathcal{I}_{night}$}
    \State Initialize $\mathcal{I}_{day-like}$ as an empty set
    \For{each image $i$ in $\mathcal{I}_{night}$}
        \State $G(i) \gets$ Apply GSAD model $G$ to $i$
        \State Add $G(i)$ to $\mathcal{I}_{day-like}$
    \EndFor
\EndProcedure

\Statex
\Procedure{PrepareDatasetForTraining}{}
    \State $\mathcal{I}_{final} \gets \mathcal{I}_{other} \cup \mathcal{I}_{day-like}$
\EndProcedure

\Statex
\Procedure{TrainAndEnsembleModels}{$\mathcal{I}_{final}$}
    \State Train multiple object detection models on $\mathcal{I}_{final}$
    \State Apply ensemble method to integrate models into $F$
\EndProcedure

\Statex
\Procedure{ApplySuperResolution}{$\mathcal{I}_{final}$}
    \State Initialize $\mathcal{I}_{sr}$ as an empty set
    \For{each image $i$ in $\mathcal{I}_{final}$}
        \State $SR(i) \gets$ Apply DAT model to $i$
        \State Add $SR(i)$ to $\mathcal{I}_{sr}$
    \EndFor
\EndProcedure

\Statex
\Procedure{GenerateDetectionResults}{$\mathcal{I}_{sr}$}
    \For{each image $i$ in $\mathcal{I}_{sr}$}
        \State $F(i) \gets$ Apply unified model $F$ to $i$
        \State Add $F(i)$ to $\mathcal{O}$
    \EndFor
    \State \textbf{return} $\mathcal{O}$
\EndProcedure

\end{algorithmic}
\end{algorithm}
\textbf{Night-to-day conversion.} The night-time images undergo a transformation process using a GSAD model, G, designed to convert night scenes into day-like scenes. This step, producing $\mathcal{I}_{day-like}$, aims to normalize the lighting conditions across the dataset, thereby reducing the variability in illumination that can negatively impact object detection performance. In this study, we utilized the GSAD model pre-trained on the LOLv2 Synthetic dataset for the experiment. Mainly because this pre-trained model demonstrates superior performance with significant improvements in brightness, contrast, and color saturation, as shown in Figure \ref{fig:lolv2_compare}.

\subsection{Detection Models for Training}

After the data preprocessing step, a unified dataset, $I_{final}$, is prepared by combining the other time-of-the-day images, $I_{other}$, with the day-like images, $I_{day-like}$. This process ensures that the training dataset exhibits a consistent lighting condition, mimicking daytime scenes and thus facilitating more effective learning by the object detection models. The models used in this study are YOLOv8x, YOLOv9e, and Co-DETR.

\vspace{0.1in}
\textbf{YOLOv8.} YOLOv8 \cite{aboah2023real}, like YOLOv5 \cite{aboah2021vision,shoman2022region,shoman2020deep}, consists of the backbone, head, and neck. Significant advancements have been made to the YOLOv8 architecture, including a complete redesign of the architecture, improved convolutional layers in the backbone, and a more advanced detection head. Because of these updates, it is now the preferred method for real-time object detection.  The model uses the Darknet-53 backbone network, which is known for its faster and more accurate performance compared to its predecessor, the YOLOv7 network \cite{wang2023yolov7,aboah2023deepsegmenter,shoman2022gc}. The YOLOv8 model uses an anchor-free detection head to predict bounding boxes. This approach is remarkably effective due to its improved feature map and convolutional network, which result in increased precision and speed.

\vspace{0.1in}
\textbf{YOLOv9} \cite{wang2024yolov9} represents a significant advancement in real-time object detection, introducing innovative techniques such as Programmable Gradient Information (PGI) and the Generalized Efficient Layer Aggregation Network (GELAN). The architecture was designed in a manner to fundamentally overcome the challenges of information loss in deep neural networks. The YOLOv9 uses innovative Reversible Functions that are central to its architecture, ensuring its high efficiency and accuracy. Furthermore, the introduction of PGI aimed at alleviating the information bottleneck problem and preserving critical data across deep network layers. This preservation allows for the generation of dependable gradients, which facilitates precise model updates and improves overall model performance.


\vspace{0.1in}
\textbf{Co-DETR.} Co-DETR\cite{zong2023detrs} stands out in its structure with a unique collaborative training scheme that tackles limitations in the DETR architecture. It achieves this by introducing auxiliary heads alongside the main decoder. These auxiliary heads train with a one-to-many object assignment approach, unlike the decoder's attention mechanism. This strengthens the encoder's ability to learn discriminative features for objects.  Furthermore, Co-DETR leverages these auxiliary heads to create more high-quality training data specifically for the decoder, improving its focus and performance.

\subsection{Robust Post-processing Strategy}
The postprocessing technique employed in this stage includes super-resolution of test images and our proposed ensemble model for detection.

\begin{figure}[H]
  \centering
  \includegraphics[width=8cm]{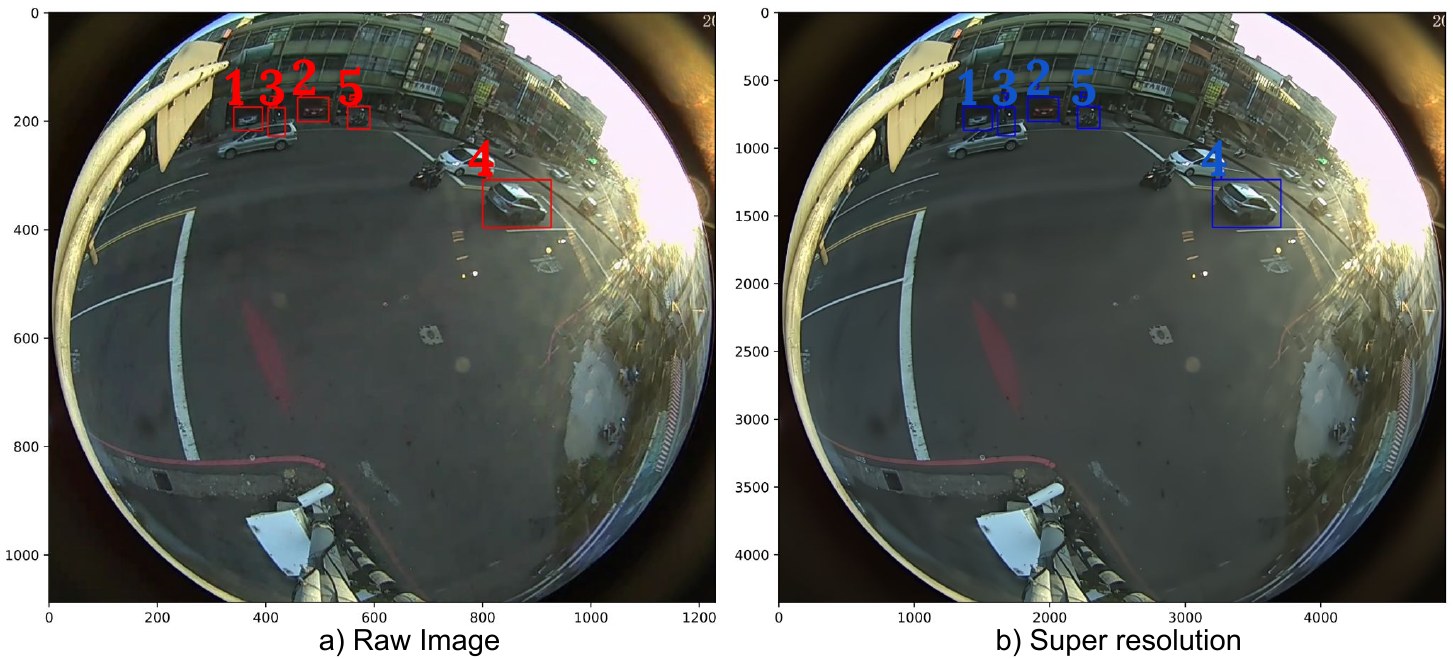}
   \caption{Using super-resolution to increase image size by a factor of four. a) raw image, b) super-resolution image}
   \label{fig:lolv3_compare}
\end{figure}

\vspace{0.1in}
\textbf{Super-resolution. } Before performing inference on our validation dataset or on the experimental test data, the images are processed through a Dual Aggregation Transformer (DAT) \cite{chen2023dual} model. This model upscales the images by a factor of four, as shown in Figure \ref{fig:lolv3_compare}, significantly enhancing the resolution and detail available for object detection.

\vspace{0.1in}

\begin{figure}[H]
  \centering
  \includegraphics[width=8cm]{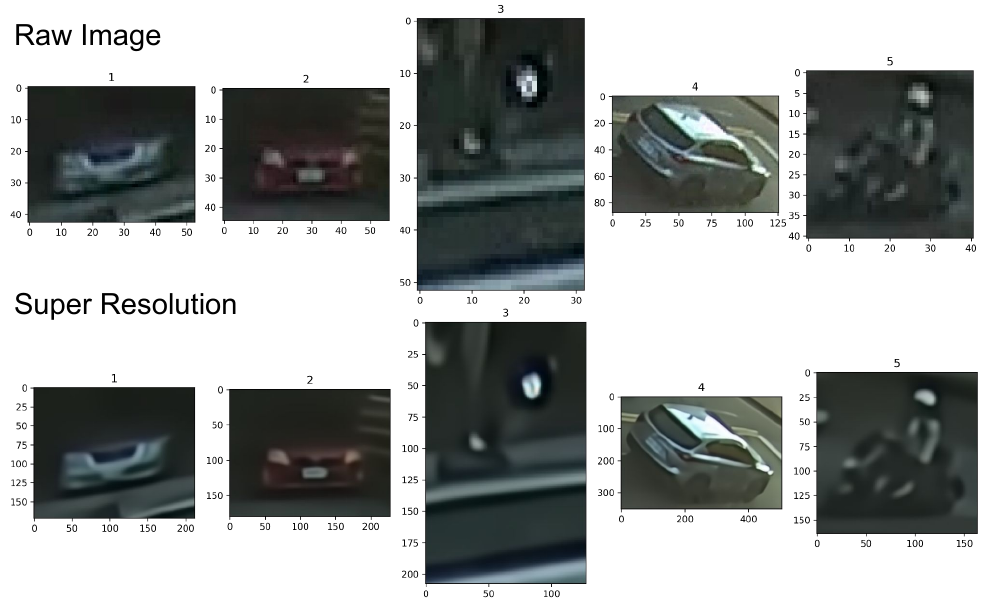}
   \caption{Using super-resolution to increase image size by a factor of four. a) raw image, b) super-resolution image}
   \label{fig:compare_object}
\end{figure}

Additionally, Figure \ref{fig:compare_object} compares the raw image with the super-resolution image at the object level. From the Figure, we observed an enhanced image quality, allowing better detection by the object detection model during inference.

\textbf{Ensemble of detection models.} The trained detection models were ensembled into one model, leveraging their respective strengths to detect various objects from the fisheye lens camera images. To achieve this, an ensemble method, weighted box fusion (WBF) \cite{solovyev2021weighted}, is then applied to integrate these models into a single, comprehensive detection model, $F$, enhancing the overall detection accuracy and robustness.

\begin{figure*}[ht]
  \centering
  \includegraphics[width=\linewidth]{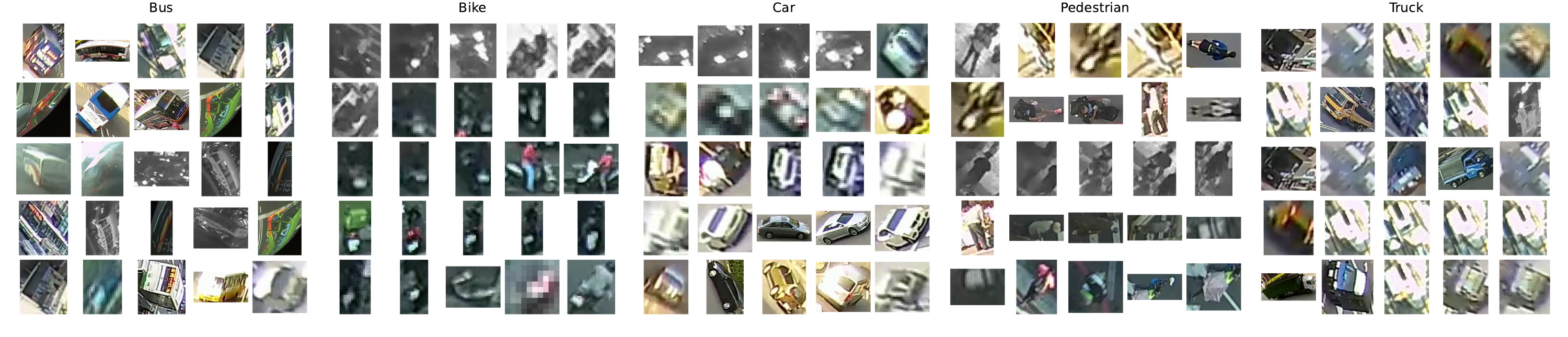}
  \caption{Sample cropped objects from the FishEye8K dataset.}
   \label{fig:vis_class}
\end{figure*}

\vspace{0.1in}
\section{Experiments}
\label{sec:exp}
\subsection{Dataset}
\textbf{Challenge of the given dataset.} As shown in Figure \ref{fig:vis_class}, the labeled objects are: \textit{Bus, Bike, Car, Pedestrian, and Truck}. The majority of the objects are labeled with small area pixels less than $64 \times 64$. Additionally, the quality of the dataset is affected by noise and blur due to the extraction process from recorded videos. Furthermore, fisheye lenses cause significant distortion, especially at the edges of the images, which can alter the appearance of objects, making it harder for detection algorithms to accurately identify and classify them. Some objects may be partially obscured by others or overlap, particularly in crowded urban scenes, making it difficult to distinguish one object from another. That is, differentiating between similar-shaped objects, such as cars and trucks or bikes and motorcycles, can be difficult, especially in low-resolution images. 

\textbf{Implementation detailed.} The proposed approach is inferenced on Intel Core i9, and NVIDIA 4090 24GB and 64GB RAM. Models are trained on Intel Xeon Silver 4210R, and 2 NVIDIA RTX A6000 48GB and 126GB RAM. Here is the list of models that are utilized for training:
\begin{itemize}
    \item YOLOv8: YOLOv8x are used for training models for comparison, multiple image scales such as 640, and 1280 are trained and validated. Other hyperparameters remained the same in the original model.  
    \item YOLOv9: At the time of writing this paper, YOLOv9e is claimed that achieved better performance compared to other YOLO variants. Therefore, the strongest YOLOv9-E with image size 1280 is used for the training model. 
    \item Co-DETR: For the large detection model like Co-DETR. The pre-trained model on foundation datasets such as COCO\cite{lin2014microsoft} or Objects365\cite{shao2019objects365} is important for improving fine-tuning model accuracy. In this research, the model is trained with image size 1024 and a fine-tuned model trained on Objects365 pre-trained + COCO dataset (Co-DETR-O365). Other hyperparameters and augmentation processes are utilized from the default model\cite{codetr2022}.
\end{itemize}

\subsection{Evaluation Metrics}
In the challenge, the evaluation of object detection models was anchored on two main metrics: the mean Average Precision (mAP) and the F1 Score. The mAP can be expressed as the mean of the AP values across all classes, where AP is calculated using the area under the precision-recall curve. The mAP can be expressed as the mean of the AP values across all classes, where AP is calculated using the area under the precision-recall curve. In a setting with 
$C$ classes, the mAP is defined as:

\begin{equation}
mAP=\frac{1}{C}\sum_{c=1}^{C}AP_{c}
\end{equation}

The F1 Score is defined as the balance mean of precision (P) and recall (R):

\begin{equation}
F1 = 2\times \frac{P\times R}{P + R}
\end{equation}

\section{Results and Discusion}

\subsection{Comparative analysis on validation set}
Table \ref{tab:compare1} provides a comparative analysis of object detection models trained and tested on the raw FishEye8K dataset. The models are evaluated based on their Average Precision (AP) from a threshold of 0.5 to 0.95, offering a rigorous metric for detection accuracy across different levels of intersection over union (IoU). 

\textbf{In detailed.} A baseline model referenced as \cite{gochoo2023fisheye8k}, which sets the initial benchmark with an AP of 0.4029. Two iterations of the YOLOv8X model with different input resolutions (640 and 1280), show APs of 0.33 and 0.36 respectively. The Co-DETR model demonstrates a competitive AP of 0.39. This comparison establishes the performance standards before the application of our proposed image enhancement techniques.

\begin{table}[ht]
  \centering
  \begin{tabular}{@{}llc@{}}
    \toprule
    ID & Method & $AP_{0.5-0.95}$  \\ 
    \midrule
    1 & Baseline \cite{gochoo2023fisheye8k} & 0.4029 \\
    2 & Yolov8X-640 \cite{Jocher_Ultralytics_YOLO_2023}  & 0.33   \\
    3 & Yolov8X-1280 \cite{Jocher_Ultralytics_YOLO_2023} & 0.36 \\
    4 & Co-DETR \cite{codetr2022} & 0.39  \\
    \bottomrule
  \end{tabular}
  \caption{Model comparison, training, and testing on raw FishEye8K dataset.}
  \label{tab:compare1}
\end{table}
In Table \ref{tab:compare2}, we extend our comparison to models trained on the image-enhanced FishEye8K dataset. The enhancements are presumed to improve model training by providing clearer and more accurate representations of objects. The models in this table include YOLOv9e-1280, achieving an AP of 0.399. The Co-DETR model, this time reaching an AP of 0.409. The Co-DETR-O365, shows a marked improvement with an AP of 0.485. Our proposed approaches, referred to as Our-1 and Our-2, surpass the others with APs of 0.486 and 0.489, respectively. The advancement of our models is indicated by the higher precision rates, demonstrating the effectiveness of our methods in handling the unique challenges of fisheye lens distortions and low-light conditions.

\begin{table}
  \centering
  \begin{tabular}{@{}llc@{}}
    \toprule
    ID & Method & $AP_{0.5-0.95}$  \\ 
    \midrule
    5 & YOLOv9e-1280 \cite{wang2024yolov9} & 0.399  \\
    6 & Co-DETR \cite{codetr2022} & 0.409  \\
    7 & Co-DETR-O365 \cite{codetr2022} & 0.485  \\
    8 & Our-1 &  0.486 \\
    \textbf{9} & \textbf{Our-2} &  \textbf{0.489}  \\
    \bottomrule
  \end{tabular}
  \caption{Model comparison, training, and testing on image enhancement FishEye8K dataset. Our-1: Co-DETR-O365 test on validation set contains day-like images. Our-2: Co-DETR-O365 train on training set contains day-like images and tests on high-resolution images.}
  \label{tab:compare2}
\end{table}

\begin{figure}[ht]
  \centering
  \includegraphics[width=\linewidth]{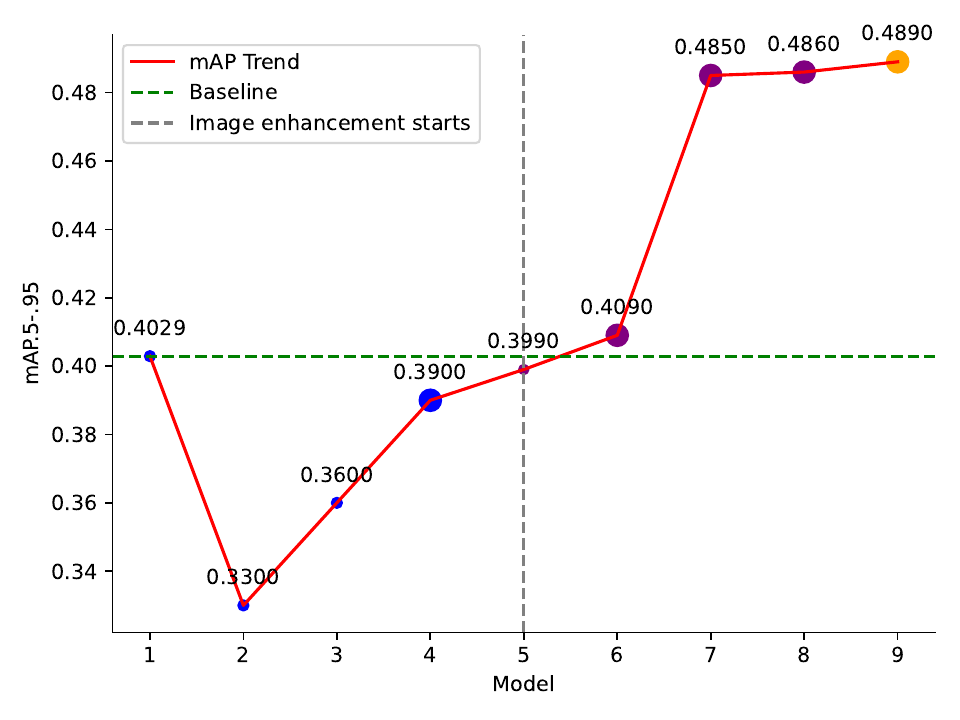}
  \caption{Model performance comparison.}
  \label{fig_compare_map}
\end{figure}

Figure \ref{fig_compare_map} illustrates the trend in model performance as measured by mAP across the range of 0.5 to 0.95 IoU. The baseline AP is indicated by a dashed green line, against which subsequent model performances are measured. The red line denotes the mAP trend, clearly showing a significant improvement in AP values with the introduction of image enhancement, marked by a dashed vertical line. Models 1 through 4, which are trained on raw images, show lower mAP values. Beginning with model 5 and extending to model 9, there is a notable uptick in mAP, underscoring the beneficial impact of image enhancement on model accuracy. Our proposed methods (Our-1 and Our-2) stand out with the highest AP values, affirming the superiority of the advancements we have implemented over existing models.


\subsection{Model performance on CVPR challenge set}

\begin{figure}[ht]
  \centering
  \includegraphics[width=\linewidth]{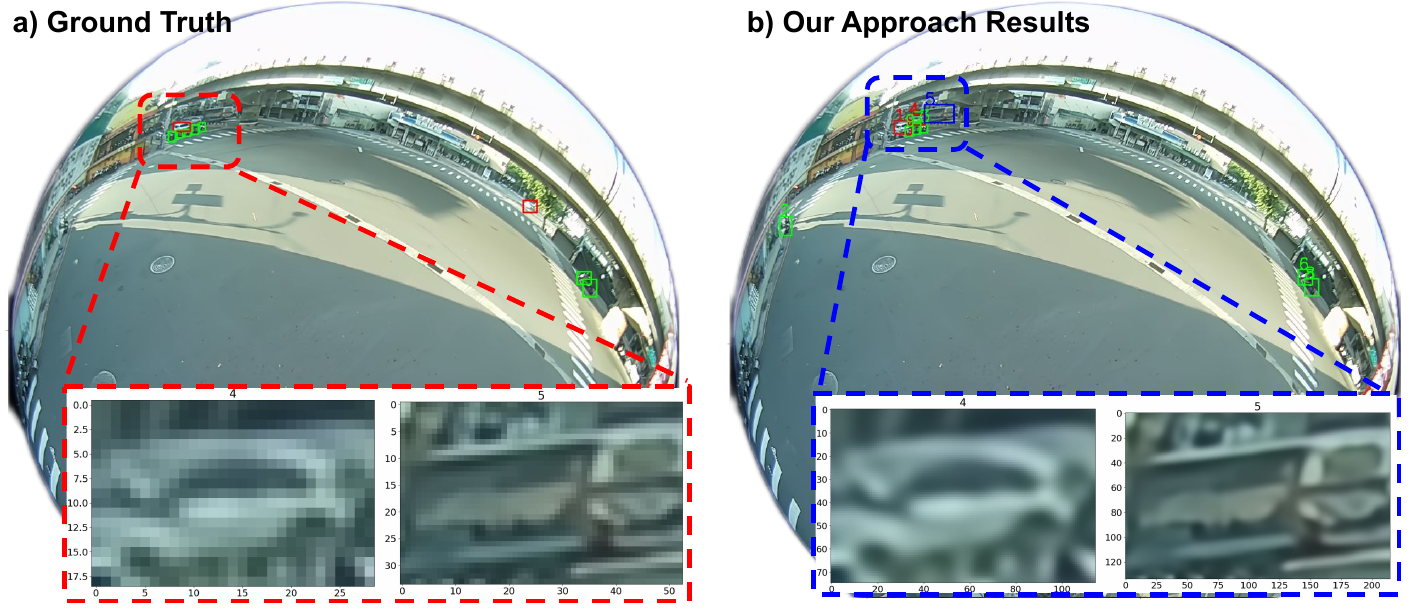}
  \caption{Qualitative Result}
  \label{fig:qualitative_resul}
\end{figure}

  
Table \ref{table:cvpr_test_compare} presents the standings from the public leaderboard of AIC24 Track 4\cite{Shuo24AIC24}, evaluating the performance of various teams on the full test set as measured by the F1 Score. In the competition, our team achieved a ranking of \textbf{5th out of 52 teams} with an F1 Score of \textbf{0.5965}. Figure \ref{fig:qualitative_resul} presents a qualitative comparison between ground truth annotations and the results achieved using our image enhancement and detection framework. Our approach not only matches the ground truth but also reveals additional objects, highlighting the potential of our method to improve resolution and detection capabilities beyond the limitations of original low-resolution annotations.

\begin{table}[h]
\caption{Public leaderboard of AIC24 Track 4 on the full test set.}
  \label{table:cvpr_test_compare}
  \centering
  \begin{tabular}{@{}cccc@{}}
    \toprule
    Rank & Team ID & Team Name & F1 Score \\ 
    \midrule
    1 & 9 & VNPT AI & 0.6406 \\ 
    2 & 40 & 	NetsPresso & 0.6196 \\
    3 & 5 & SKKU-AutoLab & 0.6194 \\
    4 & 63 & UIT-AICLUB & 	0.6077 \\
    \textbf{5} & \textbf{15} & \textbf{SKKU-NDSU (Our)} & \textbf{0.5965} \\
    6 & 33 & MCPRL & 0.5883 \\
    7 & 156 & zzl & 0.5828 \\
    8 & 52 & DeepDrivePL & 0.5825 \\
    9 & 86 & NCKU-ACVLAB & 0.5764 \\
    10 & 13 & FRDC-SH & 0.5637 \\
    \bottomrule
  \end{tabular}
  
\end{table}

\section{Conclusion}
We developed an efficient object detection system that is robust to varying lighting conditions and times of the day. To achieve this goal, we proposed a unique data processing strategy referred to as the \textbf{\textit{``low-light image enhancement framework"}} and utilized an ensemble of YOLOv8, YOLOv9, and Co-DETR as our objection detection model. Our experimental results demonstrated the effectiveness and robustness of the proposed system in detecting objects from fisheye lens camera images. 

\section*{Acknowledgement}
This research was supported by a grant [2022-MOIS38-002 (RS-2022-ND630021)] from the Ministry of Interior and Safety (MOIS)'s project for proactive technology development safety accidents for vulnerable groups and facilities, and this research was supported by a grant from the Korean Government (MSIT) to the NRF [RS-2023-00250166]. This work is financially supported by Korea Ministry of Land, Infrastructure and Transport(MOLIT) as Innovative Talent Education Program for Smart City. 

{\small
\bibliographystyle{ieee_fullname}
\bibliography{egbib}

\begin{thebibliography}{10}\itemsep=-1pt

\bibitem{abdelhamed2018high}
Abdelrahman Abdelhamed, Stephen Lin, and Michael~S Brown.
\newblock A high-quality denoising dataset for smartphone cameras.
\newblock In {\em Proceedings of the IEEE conference on computer vision and pattern recognition}, pages 1692--1700, 2018.

\bibitem{aboah2021vision}
Armstrong Aboah.
\newblock A vision-based system for traffic anomaly detection using deep learning and decision trees.
\newblock In {\em Proceedings of the IEEE/CVF Conference on Computer Vision and Pattern Recognition}, pages 4207--4212, 2021.

\bibitem{aboah2023deepsegmenter}
Armstrong Aboah, Ulas Bagci, Abdul~Rashid Mussah, Neema~Jakisa Owor, and Yaw Adu-Gyamfi.
\newblock Deepsegmenter: Temporal action localization for detecting anomalies in untrimmed naturalistic driving videos.
\newblock In {\em Proceedings of the IEEE/CVF Conference on Computer Vision and Pattern Recognition}, pages 5358--5364, 2023.

\bibitem{aboah2023real}
Armstrong Aboah, Bin Wang, Ulas Bagci, and Yaw Adu-Gyamfi.
\newblock Real-time multi-class helmet violation detection using few-shot data sampling technique and yolov8.
\newblock In {\em Proceedings of the IEEE/CVF conference on computer vision and pattern recognition}, pages 5349--5357, 2023.

\bibitem{agorku2023real}
Geoffery Agorku, Divine Agbobli, Vuban Chowdhury, Kwadwo Amankwah-Nkyi, Adedolapo Ogungbire, Portia~Ankamah Lartey, and Armstrong Aboah.
\newblock Real-time helmet violation detection using yolov5 and ensemble learning.
\newblock {\em arXiv preprint arXiv:2304.09246}, 2023.

\bibitem{al2021improved}
Mohammed~AA Al-qaness, Aaqif~Afzaal Abbasi, Hong Fan, Rehab~Ali Ibrahim, Saeed~H Alsamhi, and Ammar Hawbani.
\newblock An improved yolo-based road traffic monitoring system.
\newblock {\em Computing}, 103:211--230, 2021.

\bibitem{ardianto2023fast}
Sandy Ardianto, Hsueh-Ming Hang, and Wen-Huang Cheng.
\newblock Fast vehicle detection and tracking on fisheye traffic monitoring video using motion trail.
\newblock In {\em 2023 IEEE International Symposium on Circuits and Systems (ISCAS)}, pages 1--5. IEEE, 2023.

\bibitem{chen2022simple}
Liangyu Chen, Xiaojie Chu, Xiangyu Zhang, and Jian Sun.
\newblock Simple baselines for image restoration.
\newblock {\em arXiv preprint arXiv:2204.04676}, 2022.

\bibitem{chen2023dual}
Zheng Chen, Yulun Zhang, Jinjin Gu, Linghe Kong, Xiaokang Yang, and Fisher Yu.
\newblock Dual aggregation transformer for image super-resolution.
\newblock In {\em ICCV}, 2023.

\bibitem{chu2022nafssr}
Xiaojie Chu, Liangyu Chen, and Wenqing Yu.
\newblock Nafssr: Stereo image super-resolution using nafnet.
\newblock In {\em Proceedings of the IEEE/CVF Conference on Computer Vision and Pattern Recognition}, pages 1239--1248, 2022.

\bibitem{fedorov2019traffic}
Aleksandr Fedorov, Kseniia Nikolskaia, Sergey Ivanov, Vladimir Shepelev, and Alexey Minbaleev.
\newblock Traffic flow estimation with data from a video surveillance camera.
\newblock {\em Journal of Big Data}, 6:1--15, 2019.

\bibitem{gavrila2000pedestrian}
Dariu~M Gavrila.
\newblock Pedestrian detection from a moving vehicle.
\newblock In {\em Computer Vision—ECCV 2000: 6th European Conference on Computer Vision Dublin, Ireland, June 26--July 1, 2000 Proceedings, Part II 6}, pages 37--49. Springer, 2000.

\bibitem{gochoo2023fisheye8k}
Munkhjargal Gochoo, Munkh-Erdene Otgonbold, Erkhembayar Ganbold, Jun-Wei Hsieh, Ming-Ching Chang, Ping-Yang Chen, Byambaa Dorj, Hamad Al~Jassmi, Ganzorig Batnasan, Fady Alnajjar, et~al.
\newblock Fisheye8k: A benchmark and dataset for fisheye camera object detection.
\newblock In {\em Proceedings of the IEEE/CVF Conference on Computer Vision and Pattern Recognition}, pages 5304--5312, 2023.

\bibitem{hou2024global}
Jinhui Hou, Zhiyu Zhu, Junhui Hou, Hui Liu, Huanqiang Zeng, and Hui Yuan.
\newblock Global structure-aware diffusion process for low-light image enhancement.
\newblock {\em Advances in Neural Information Processing Systems}, 36, 2024.

\bibitem{jeon2023leveraging}
Yuntae Jeon, Dai~Quoc Tran, Minsoo Park, and Seunghee Park.
\newblock Leveraging future trajectory prediction for multi-camera people tracking.
\newblock In {\em Proceedings of the IEEE/CVF Conference on Computer Vision and Pattern Recognition}, pages 5398--5407, 2023.

\bibitem{Jocher_Ultralytics_YOLO_2023}
Glenn Jocher, Ayush Chaurasia, and Jing Qiu.
\newblock {Ultralytics YOLO}, 2023.

\bibitem{lin2014microsoft}
Tsung-Yi Lin, Michael Maire, Serge Belongie, James Hays, Pietro Perona, Deva Ramanan, Piotr Doll{\'a}r, and C~Lawrence Zitnick.
\newblock Microsoft coco: Common objects in context.
\newblock In {\em Computer Vision--ECCV 2014: 13th European Conference, Zurich, Switzerland, September 6-12, 2014, Proceedings, Part V 13}, pages 740--755. Springer, 2014.

\bibitem{madhavi2023traffic}
G~Bindu Madhavi, A~Durga Bhavani, Y~Sowmya Reddy, Ajmeera Kiran, N~Thulasi Chitra, and Pundru Chandra~Shaker Reddy.
\newblock Traffic congestion detection from surveillance videos using deep learning.
\newblock In {\em 2023 International Conference on Computer, Electronics \& Electrical Engineering \& their Applications (IC2E3)}, pages 1--5. IEEE, 2023.

\bibitem{mandal2020artificial}
Vishal Mandal, Abdul~Rashid Mussah, Peng Jin, and Yaw Adu-Gyamfi.
\newblock Artificial intelligence-enabled traffic monitoring system.
\newblock {\em Sustainability}, 12(21):9177, 2020.

\bibitem{nah2019ntire}
Seungjun Nah, Sungyong Baik, Seokil Hong, Gyeongsik Moon, Sanghyun Son, Radu Timofte, and Kyoung Mu~Lee.
\newblock Ntire 2019 challenge on video deblurring and super-resolution: Dataset and study.
\newblock In {\em Proceedings of the IEEE/CVF conference on computer vision and pattern recognition workshops}, pages 0--0, 2019.

\bibitem{nah2017deep}
Seungjun Nah, Tae Hyun~Kim, and Kyoung Mu~Lee.
\newblock Deep multi-scale convolutional neural network for dynamic scene deblurring.
\newblock In {\em Proceedings of the IEEE conference on computer vision and pattern recognition}, pages 3883--3891, 2017.

\bibitem{nguyen2014object}
Thao Nguyen, Eun-Ae Park, Jiho Han, Dong-Chul Park, and Soo-Young Min.
\newblock Object detection using scale invariant feature transform.
\newblock In {\em Genetic and Evolutionary Computing: Proceedings of the Seventh International Conference on Genetic and Evolutionary Computing, ICGEC 2013, August 25-27, 2013-Prague, Czech Republic}, pages 65--72. Springer, 2014.

\bibitem{qiu2021deep}
Linrun Qiu, Dongbo Zhang, Yuan Tian, and Najla Al-Nabhan.
\newblock Deep learning-based algorithm for vehicle detection in intelligent transportation systems.
\newblock {\em The Journal of Supercomputing}, 77(10):11083--11098, 2021.

\bibitem{redmon2016you}
Joseph Redmon, Santosh Divvala, Ross Girshick, and Ali Farhadi.
\newblock You only look once: Unified, real-time object detection.
\newblock In {\em Proceedings of the IEEE conference on computer vision and pattern recognition}, pages 779--788, 2016.

\bibitem{sarrab2020development}
Mohammed Sarrab, Supriya Pulparambil, and Medhat Awadalla.
\newblock Development of an iot based real-time traffic monitoring system for city governance.
\newblock {\em Global Transitions}, 2:230--245, 2020.

\bibitem{shao2019objects365}
Shuai Shao, Zeming Li, Tianyuan Zhang, Chao Peng, Gang Yu, Xiangyu Zhang, Jing Li, and Jian Sun.
\newblock Objects365: A large-scale, high-quality dataset for object detection.
\newblock In {\em Proceedings of the IEEE/CVF international conference on computer vision}, pages 8430--8439, 2019.

\bibitem{shirpour2021traffic}
Mohsen Shirpour, Nima Khairdoost, Michael~A Bauer, and Steven~S Beauchemin.
\newblock Traffic object detection and recognition based on the attentional visual field of drivers.
\newblock {\em IEEE Transactions on Intelligent Vehicles}, 8(1):594--604, 2021.

\bibitem{shoman2020deep}
Maged Shoman, Armstrong Aboah, and Yaw Adu-Gyamfi.
\newblock Deep learning framework for predicting bus delays on multiple routes using heterogenous datasets.
\newblock {\em Journal of Big Data Analytics in Transportation}, 2:275--290, 2020.

\bibitem{shoman2022gc}
Maged Shoman, Armstrong Aboah, Abdulateef Daud, and Yaw Adu-Gyamfi.
\newblock Gc-gru-n for traffic prediction using loop detector data.
\newblock {\em arXiv preprint arXiv:2211.08541}, 2022.

\bibitem{shoman2022region}
Maged Shoman, Armstrong Aboah, Alex Morehead, Ye Duan, Abdulateef Daud, and Yaw Adu-Gyamfi.
\newblock A region-based deep learning approach to automated retail checkout.
\newblock In {\em Proceedings of the IEEE/CVF Conference on Computer Vision and Pattern Recognition}, pages 3210--3215, 2022.

\bibitem{shoman2022multi}
Maged Shoman, Mark Amo-Boateng, and Yaw Adu-Gyamfi.
\newblock Multi-purpose, multi-step deep learning framework for network-level traffic flow prediction.
\newblock {\em Advances in Data Science and Adaptive Analysis}, 14(03n04):2250010, 2022.

\bibitem{shoman}
Maged Shoman, Dongdong Wang, Armstrong Aboah, and Mohamed Abdel-Aty.
\newblock Enhancing traffic safety with parallel dense video captioning for end-to-end event analysis.
\newblock June 2024.

\bibitem{shou2022object}
Yuntao Shou, Tao Meng, Wei Ai, Canhao Xie, Haiyan Liu, and Yina Wang.
\newblock Object detection in medical images based on hierarchical transformer and mask mechanism.
\newblock {\em Computational Intelligence and Neuroscience}, 2022, 2022.

\bibitem{solovyev2021weighted}
Roman Solovyev, Weimin Wang, and Tatiana Gabruseva.
\newblock Weighted boxes fusion: Ensembling boxes from different object detection models.
\newblock {\em Image and Vision Computing}, pages 1--6, 2021.

\bibitem{soltanikazemi2023real}
Elham Soltanikazemi, Ashwin Dhakal, Bijaya~Kumar Hatuwal, Imad~Eddine Toubal, Armstrong Aboah, and Kannappan Palaniappan.
\newblock Real-time helmet violation detection in ai city challenge 2023 with genetic algorithm-enhanced yolov5.
\newblock In {\em 2023 IEEE Applied Imagery Pattern Recognition Workshop (AIPR)}, pages i--x. IEEE, 2023.

\bibitem{song2022extendable}
Hwanjun Song, Deqing Sun, Sanghyuk Chun, Varun Jampani, Dongyoon Han, Byeongho Heo, Wonjae Kim, and Ming-Hsuan Yang.
\newblock An extendable, efficient and effective transformer-based object detector.
\newblock {\em arXiv preprint arXiv:2204.07962}, 2022.

\bibitem{stuparu2020vehicle}
Delia-Georgiana Stuparu, Radu-Ioan Ciobanu, and Ciprian Dobre.
\newblock Vehicle detection in overhead satellite images using a one-stage object detection model.
\newblock {\em Sensors}, 20(22):6485, 2020.

\bibitem{9801825}
Dai~Quoc Tran, Minsoo Park, Yuntae Jeon, Jinyeong Bak, and Seunghee Park.
\newblock Forest-fire response system using deep-learning-based approaches with cctv images and weather data.
\newblock {\em IEEE Access}, 10:66061--66071, 2022.

\bibitem{tran2020damage}
Dai~Quoc Tran, Minsoo Park, Daekyo Jung, and Seunghee Park.
\newblock Damage-map estimation using uav images and deep learning algorithms for disaster management system.
\newblock {\em Remote Sensing}, 12(24):4169, 2020.

\bibitem{vaidwan2021study}
Hritik Vaidwan, Nikhil Seth, Anil~Singh Parihar, and Kavinder Singh.
\newblock A study on transformer-based object detection.
\newblock In {\em 2021 international conference on intelligent technologies (CONIT)}, pages 1--6. IEEE, 2021.

\bibitem{wang2023yolov7}
Chien-Yao Wang, Alexey Bochkovskiy, and Hong-Yuan~Mark Liao.
\newblock Yolov7: Trainable bag-of-freebies sets new state-of-the-art for real-time object detectors.
\newblock In {\em Proceedings of the IEEE/CVF conference on computer vision and pattern recognition}, pages 7464--7475, 2023.

\bibitem{wang2024yolov9}
Chien-Yao Wang, I-Hau Yeh, and Hong-Yuan~Mark Liao.
\newblock Yolov9: Learning what you want to learn using programmable gradient information.
\newblock {\em arXiv preprint arXiv:2402.13616}, 2024.

\bibitem{wang2014hybrid}
Huan Wang and Haichuan Zhang.
\newblock A hybrid method of vehicle detection based on computer vision for intelligent transportation system.
\newblock {\em International Journal of Multimedia and Ubiquitous Engineering}, 9(6):105--118, 2014.

\bibitem{wang2010boosting}
Jian-Gang Wang, Jun Li, Wei-Yun Yau, and Eric Sung.
\newblock Boosting dense sift descriptors and shape contexts of face images for gender recognition.
\newblock In {\em 2010 IEEE computer society conference on computer vision and pattern recognition-workshops}, pages 96--102. IEEE, 2010.

\bibitem{Shuo24AIC24}
Shuo Wang, David~C. Anastasiu, Zheng Tang, Ming-Ching Chang, Yue Yao, Liang Zheng, Mohammed~Shaiqur Rahman, Meenakshi~S. Arya, Anuj Sharma, Pranamesh Chakraborty, Sanjita Prajapati, Quan Kong, Norimasa Kobori, Munkhjargal Gochoo, Munkh-Erdene Otgonbold, Ganzorig Batnasan, Fady Alnajjar, Ping-Yang Chen, Jun-Wei Hsieh, Xunlei Wu, Sameer~Satish Pusegaonkar, Yizhou Wang, Sujit Biswas, and Rama Chellappa.
\newblock The 8th {AI City Challenge}.
\newblock In {\em The IEEE Conference on Computer Vision and Pattern Recognition (CVPR) Workshops}, June 2024.

\bibitem{yadav2023video}
Divakar Yadav, Arti Jain, Saumya Asati, and Arun~Kumar Yadav.
\newblock Video anomaly detection for pedestrian surveillance.
\newblock In {\em Computer Vision and Machine Intelligence: Proceedings of CVMI 2022}, pages 489--500. Springer, 2023.

\bibitem{ye2020autonomous}
Tao Ye, Zhihao Zhang, Xi Zhang, and Fuqiang Zhou.
\newblock Autonomous railway traffic object detection using feature-enhanced single-shot detector.
\newblock {\em IEEE Access}, 8:145182--145193, 2020.

\bibitem{codetr2022}
Zhuofan Zong, Guanglu Song, and Yu Liu.
\newblock Detrs with collaborative hybrid assignments training, 2022.

\bibitem{zong2023detrs}
Zhuofan Zong, Guanglu Song, and Yu Liu.
\newblock Detrs with collaborative hybrid assignments training.
\newblock In {\em Proceedings of the IEEE/CVF international conference on computer vision}, pages 6748--6758, 2023.

\end{thebibliography}
}

\end{document}